\pdfoutput=1

\documentclass[11pt]{article}

\usepackage[]{acl}

\usepackage{times}
\usepackage{latexsym}
\usepackage{graphicx}
\usepackage{multirow}
\usepackage{amssymb}
\usepackage{bbding}
\usepackage{amsmath}
\usepackage{mathrsfs}
\usepackage{hyperref}
\usepackage{caption}

\usepackage{times}
\usepackage{latexsym}
\usepackage{graphicx}
\usepackage{booktabs} 
\usepackage[export]{adjustbox}
\usepackage{graphicx}

\usepackage{inconsolata}
\usepackage{microtype}
\usepackage{graphicx}
\usepackage{multirow}
\usepackage{subcaption}
\usepackage{amsmath, amssymb, mathtools}
\usepackage[export]{adjustbox}
\usepackage{hyperref}

\usepackage[T1]{fontenc}

\usepackage[utf8]{inputenc}

\usepackage{microtype}

\title{Aqulia-Med LLM: Pioneering Full-Process Open-Source Medical Language Models}

\author{Lulu Zhao$^{1}$, Weihao Zeng$^{2}$\thanks{\ \ \ Work done during the internship at BAAI.}, Xiaofeng Shi$^{1}$, Hua Zhou$^{1}$, Donglin Hao$^{1}$, Yonghua Lin$^{1}$\\
$^1$Beijing Academy of Artificial Intelligence, BAAI, Beijing, China \\
$^2$Beijing University of Posts and Telecommunications, Beijing, China\\
\texttt{\{llzhao,xfshi,zhouhua,dlhao,yhlin\}@baai.ac.cn}\\
   \texttt{\{ZengWH\}@bupt.edu.cn}
  }

\begin{document}
\maketitle
\begin{abstract}
Recently, both closed-source LLMs and open-source communities have made significant strides, outperforming humans in various general domains. However, their performance in specific professional fields such as medicine, especially within the open-source community, remains suboptimal due to the complexity of medical knowledge. We propose Aquila-Med, a bilingual medical LLM based on Aquila, addressing these challenges through continue pre-training, supervised fine-tuning (SFT), and reinforcement learning from human feedback (RLHF). We construct a large-scale Chinese and English medical dataset for continue pre-training and a high-quality SFT dataset, covering extensive medical specialties. Additionally, we develop a high-quality Direct Preference Optimization (DPO) dataset for further alignment. Aquila-Med achieves notable results across single-turn, multi-turn dialogues, and medical multiple-choice questions, demonstrating the effectiveness of our approach. We open-source the datasets and the entire training process, contributing valuable resources to the research community. Our models and datasets will released at https://huggingface.co/BAAI/AquilaMed-RL. 
\end{abstract}

\section{Introduction}
Recently, both closed source LLMs \cite{achiam2023gpt} and open source communities \cite{touvron2023llama1, touvron2023llama} have made great progress and surpassed humans in a range of general areas. However, they have not performed very well in specific professional fields such as medicine, especially for the open source community \cite{labrak2024biomistral,han2023medalpaca,yang2024advancing}. This is because the complex and specialized medical domain knowledge is a great challenge to successfully develop an accurate and safe medical LLM \cite{singhal2022large}. We believe that medical LLMs have great application potential and can be valuable in diagnostic assistance, consultation, drug recommendation, etc. As of now, there are some medical LLMs in this field, but these works rely entirely on SFT training \cite{zhang-etal-2023-huatuogpt,zhang2024ultramedical}. As we all known, pre-training is a key stage in learning domain knowledge \cite{zeng2023futuretod,zeng2024divtod}, and relying only on SFT will cause the model to only give answers in a fixed format. For dataset, most of them only concern on the data construction of the SFT stage \cite{yang2023zhongjing,zhang-etal-2023-huatuogpt} or pay attention to single-turn dialogues \cite{li2023chatdoctor,zhang2023alpacare,li2023huatuo26m, tian2023chimedgpt}, ignoring the scenarios of multi-turn interactions in real doctor-patient dialogues. In addition, the training datasets are all monolingual and only contain dialogue-type QA data.

To solve the above issues, we propose a bilingual medical LLM based on Aquila\footnote{https://github.com/FlagAI-Open/Aquila2}, namely Aquila-Med, which implements the entire process from continue pre-training, SFT to RLHF. In addition, for continue pre-trained, we build a large-scale Chinese and English medical dataset. A high-quality Chinese and English medical SFT dataset is also constructed, comprising about approximately 330,000 examples, covering 15+ departments and 100+ disease specialties, and we also construct 13,000 high-quality DPO pairs, which include various forms such as QA and medical multiple-choice questions. It is worth noting that we are the first one to open-source the construction process of the three datasets and the entire training process. These high-quality SFT and DPO datasets will also be open-sourced to help more researchers in the open source community.

Specifically, we first collect a large amount of real medical corpus, which comes from medical data classified from massive pre-training data for Aquila, open source SFT synthetic data, and a certain proportion of general data. We then do a continue pre-training based on the Aquila to obtain a base model with a medical foundation. Secondly, we collect a large amount of open source SFT medical data, and use a variety of data selection methods to filter the quality of single-turn dialogues and multi-turn dialogues respectively. Our high-quality medical SFT dataset includes: single-turn Chinese medical dialogue data, single-turn English medical dialogue data, multi-turn Chinese medical dialogue data, and medical subject knowledge multiple-choice questions, with the aim of enhancing the model's understanding and generalization capabilities in the medical domain. It is worth noting that the dataset here is partly derived from real-world medical diagnosis dialogues and partly from the construction of GPT-3.5. We hope that the model can not only generate informative, clear and logical responses, but also give more professional and personalized consultations like doctors. In the RLHF stage, based on the results of SFT, we used GPT-4 to construct a positive-negative medical data pairs. Finally, we use the Direct Preference Optimization (DPO) \cite{rafailov2023direct} algorithm to align the output of the model with the human expression style.

\begin{figure*}[t]
\centering
\includegraphics[width=16cm, height=7.5cm]{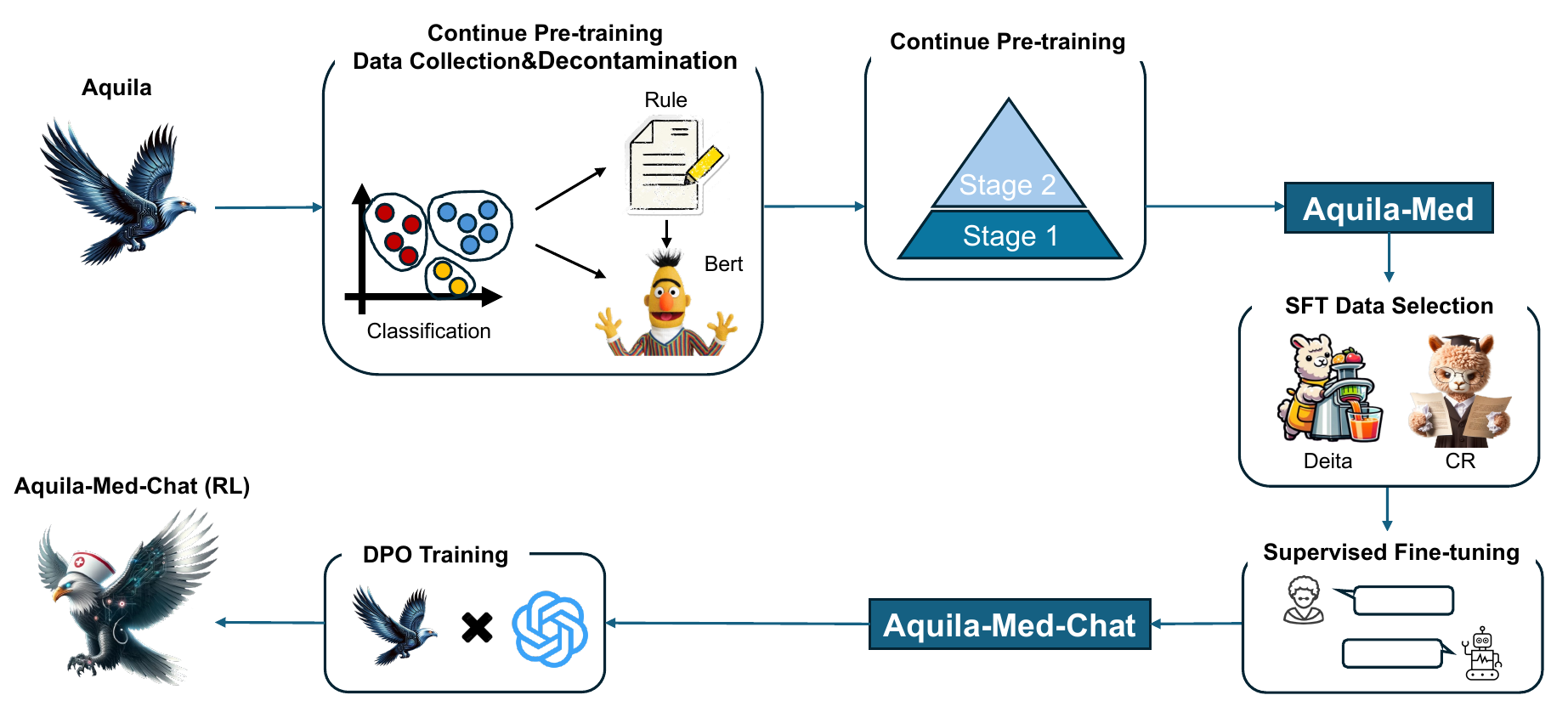}
\caption{The overall pipline of Aquila-Med-Chat (RL), which includes the continue pre-training, supervised fine-tuning, and the DPO process.}
\label{fig:pipline}
\vspace{-0.7cm}
\end{figure*} 

After extensive training and optimization, we successfully developed Aquila-Med. We also comprehensively evaluated the common benchmarks in the medical field, covering single-turn dialogue, multi-turn dialogue, and medical multiple-choice questions, involving four capability dimensions. The experimental results show that our model achieves good performance, proving that our proposed datasets effectively enhance the model's ability to handle single-turn and multi-turn medical consultations.

The main contributions of this paper are as follows: (1) We are the first to implement a full-process from pre-training, SFT to RLHF for a Chinese and English medical LLM, Aquila-Med. (2) We are the first to introduce the construction process of three datasets in the medical domain in detail: pre-training, SFT and DPO. We will make the SFT and DPO datasets public. (3) We conduct experiments on multiple Chinese and English benchmarks to verify the effectiveness and reliability of our proposed datasets.

\section{Methodology}
In this section, we introduce three stages of model training: continue pre-training, SFT, and RLHF. Each stage includes the data construction process. Each step is discussed sequentially to reflect the research workflow. The whole process is shown in Figure \ref{fig:pipline}.

\subsection{Continue Pre-training}
\subsubsection{Data Collection and Decontamination}
In this section, we will outline the process of building Aquila-Med-cpt from a massive general pre-training database, including how to collect medical-related corpora from pre-training databases, rule-based quality filtering methods, and LLM-based data quality selection methods. It is worth noting that this method is also applicable to any domain.

\noindent\textbf{Data Classification}\quad Since Aquila's general pre-training (Aquila-pt) corpus comes from multiple data sources, it already contains domain information. However, since there is no clear domain label, we first need to classify the data to make full use of the medical domain data in Aquila-pt. Specifically, we first randomly sample 20k data from Aquila-pt and use the upsampling method to ensure that the ratio of Chinese and English being 1:1. Based on the sampled data, GPT-4 is used to perform two rounds of domain label annotation to improve label accuracy. The data with different labels twice are removed, and finally 17k seed data is retained. Then we design a classifier using the Bert-based multilingual pre-training model. The parameter settings are as follows: batch-size is 64, learning rate is 2e-5, training epoch is 10, and the optimal checkpoint is selected according to the accuracy. The medical domain F1 of the classifier can reach 84\%.

\noindent\textbf{Rule-based Data Quality Filtering}\quad Since Aquila-pt mostly comes from web pages, the overall quality is not high. In order to remove the noise data, we design a rule-based data filtering solution, including rules for removing data with insufficient tokens, excessive special characters, toxic content, and private information.

\noindent\textbf{LLM-based Data Quality Filtering}\quad By sampling and checking the data after rule filtering, we found that there exists the following problems: (1) the data contains advertising and marketing information, which will greatly affect the output preference of the trained model; (2) the data contains grammatical errors, semantic incoherence, splicing of multiple unrelated content, image and video editing information, etc. We believe that such data is not beneficial for model training because the model cannot obtain much valuable information through autoregressive learning. Therefore, we design a quality scoring regression model based on LLM to score data quality and further filter out low-quality content. Specifically, we extract 20k data from the rule-based filtered data, score them twice using the GPT4, ranging from 0 to 6, and remove the data with a difference of about 2 points between the two scores, and finally obtained 15k training data. Then we train a scoring model based on the Bert multilingual pre-trained model, using batch-size of 128, learning rate of 3e-4, and train epoch of 10. We set a threshold for high-quality data filtering.


\subsubsection{Training Strategy}
Our domain pre-training is divided into two stages. The Stage 1 is the training of ordinary quality domain data, and the Stage 2 is the training of high-quality domain data.

\noindent\textbf{Stage 1:}\quad The aim is to prevent the model capability from being significantly degraded due to the large difference between pre-training and continue pre-training data. We use medical domian data filtered by rules and general data with a certain ratio. The data amount is about 60B tokens.

\noindent\textbf{Stage 2:}\quad The aim is to further improve the capability of the medical domain model. We use medical domian data filtered by LLM quality model and open source medical SFT synthetic data. The data amount is about 20B tokens.

\subsubsection{Training Details}
Our model is based on Aquila-7B, which a general Chinese-English LLM with 7 billion parameters. It has been pre-trained autoregressively with 3.6T tokens. The vocabulary size of the model is 15k, the model contains 32 layers of transformers, the maximum length is 4096, the hidden layer dimension of each layer of transformer is 4096, the FFN linear layer dimension is 14336, and the GQA structure is used in attention layers, with 8 groups and 32 heads.   

For the first stage of continuous pre-training, we train on 3*8 NVIDIA A100-40G GPUs, using a batch-size of 768, a learning rate of 1e-4, a maximum length of 4096, a cosine learning rate scheduler, a warmup-ratio of 0.05, and train for one epoch. For the second stage, keeping other settings unchanged, we reduce batch-size to 384, learning rate to 1e-5, and reduce warmup-ratio to 0.01. We also train for one epoch.


\subsection{Supervised Fine-Tuning}
To improve the ability of language models to engage in natural conversation, we firstly carry out SFT, which finetunes a pretrained LLM on chat-style data, including both queries and responses. In the following sections, we will delve into the details of data construction and training methods.

\subsubsection{Data Construction}
Our SFT dataset comprises a variety of question types, including medical exam multiple-choice questions, single-turn disease diagnosis, multi-turn health consultation, etc. It comes from 6 publicly available datasets, namely Chinese Medical Dialogue Data \footnote{https://github.com/Toyhom/Chinese-medical-dialogue-data}, Huatuo26M \cite{li2023huatuo26m}, MedDialog \cite{zeng-etal-2020-meddialog}, ChatMed Consult Dataset \cite{tian2023chimedgpt}, CMB-exam\footnote{https://github.com/FreedomIntelligence/CMB}, and ChatDoctor \cite{li2023chatdoctor}. These datasets contain not only real doctor-patient dialogues, but also dialogues generated from GPT-3.5. We believe this ensures the diversity of the dataset. 

Since a relatively small high-quality dataset has been shown to be sufficient for fine-tuning LLM, we focus on how to automatically filter "good data" from massive data to ensure competitive performance with a minimal amount of data. Similar to common data cleaning operations, we first remove duplicates and data related to security issues such as violence, bias, and pornography. In the following sections, we specifically introduce the data filtering methods.

\noindent\textbf{Single-turn Medical Dialogue Data} \quad Following \citet{liu2024what, zeng2024automatic}, we believe that "good data" should have a complex instruction and a high-quality response. Therefore, We adopt the approach from Deita \cite{liu2024what}, which employs a complexity model and a quality model to score each instance along two dimensions: instruction complexity and response quality. The complexity model assigns a complexity score $c_i$ to each instance, while the quality model assigns a quality score $q_i$, reflecting the quality of the response.
By multiplying $c_i$ with $q_i$, we combine the complexity score and quality score to obtain a comprehensive score, that is, $s_i=c_i*q_i$. Finally, we set a score threshold to select the most effective data instances in the massive data pool.

\noindent\textbf{Multi-turn Medical Dialogue Data} \quad For multi-turn dialogues, we first use Deita to calculate the score $s_i$ of each turn separately, and average them to obtain the final score of the entire dialogue. However, we found that there are two special problems in multi-turn dialogues compared to single-turn dialogues: (1) The correlation between different turn is very low, resulting in a negative impact of the previous information on the following; (2) The correlation between different turns is too high, resulting in a large degree of context duplication and information is redundant. Therefore, we propose a Context Relevance (CR) score, which is a metric that relies on cross-entropy loss to evaluate the impact of historical information on each turn. The details are as follows:

In the instruction-tuning process, the loss of a sample pair $(H, T)$ is calculated by continuously predicting the next tokens in the current turn $T$ given their previous tokens and the history information $H$:

\begin{equation}
\begin{aligned}
L_{\theta}(t_{i}|H) = -\frac{1}{N}\sum_{i=1}^{N}logP(w_{i}^j|H,w_{i}^1,w_{i}^2,...,w_{i}^{j-1};\theta)
\end{aligned}
\end{equation}
where $H=\{t_1,t_2,...t_{i-1}\}$, $t_i$ is the current turn, $w_{i}^{j}$ is the $j$-th token in the $i$-th turn, and $N$ is the number of tokens of the current turn. We define $L_{\theta}(t_{i}|H)$ as the Conditioned Information Score, which measures the ability to generate the current turn under the guidance of corresponding historical information. 

To measure the ability of LLM to generate this turn alone, we also define a Direct Information Score:
\begin{equation}
\begin{aligned}
L_{\theta}(t_{i}) = -\frac{1}{N}\sum_{i=1}^{N}logP(w_{i}^j|w_{i}^1,w_{i}^2,...,w_{i}^{j-1};\theta)
\end{aligned}
\end{equation}
We believe that the higher Direct Information Score may indicate that the turn is more challenging or complex. Finally, we try to estimate the CR score by calculating the ratio between $L_{\theta}(t_{i})$ and $L_{\theta}(t_{i}|H)$.
\begin{equation}
\begin{aligned}
CR_{\theta}(H,T)=\frac{L_{\theta}(t_{i}|H)}{L_{\theta}(t_{i})}
\end{aligned}
\end{equation}
Here, if $r>1$, it means that historical information has a negative impact on current turn, that is, the correlation between contexts is very low. If $r<1$, it means that historical information has a positive impact on current turn, that is, the correlation between contexts is high. However, too small $r$ means that the context is highly repeated and the information is highly redundant. We also set a threshold to filter the data.

\subsubsection{Training Details}
Our model is based on Aquila-Med and the training process has the following hyperparameters: sequence length set to 2048, batch size set to 128, and peak learning rate set to 2e-6 with cosine learning rate scheduler. To prevent overfitting, weight decay of 0.1 is applied and dropout is set to 0.1. Training is parallelized on 8 A100-40G NVIDIA GPUs using the AdamW optimizer with bf16 precision and ZeRO-3. We reserve 10\% of the training set for validation and get the best checkpoint after 2 epochs.

\begin{figure}[t]
\centering
\includegraphics[width=7.7cm, height=6cm]{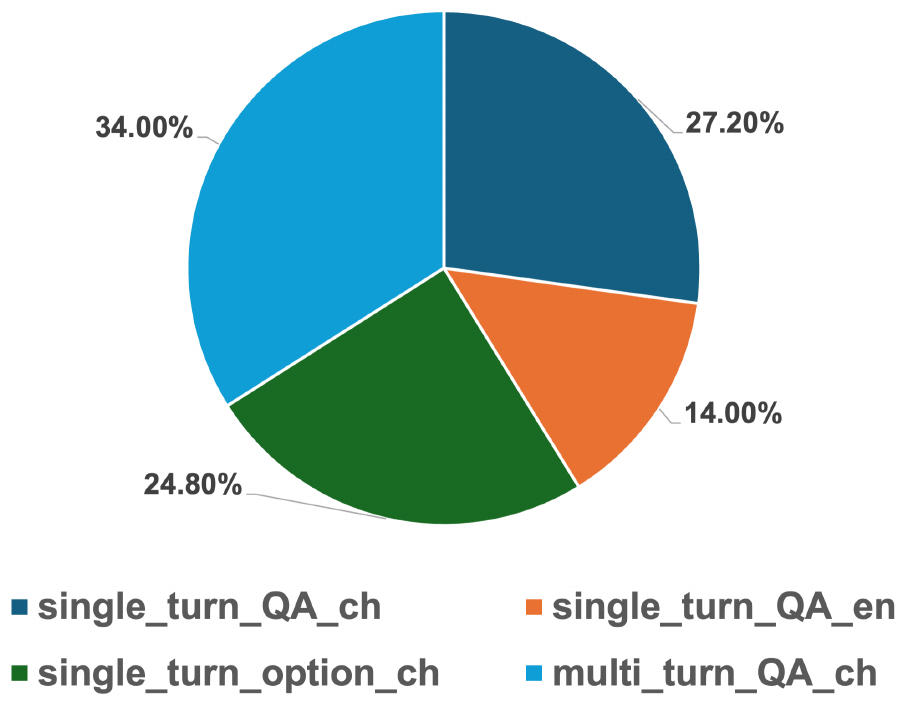}
\caption{Statistics on the distribution of our proposed SFT dataset.}
\label{fig:sft_data_distribution}
\vspace{-0.5cm}
\end{figure} 

\subsubsection{Dataset Statistics}
Through the above data filtering methods, we select s 320,000 high-quality SFT medical dataset from 199,000 instances, in which the ratio of Chinese and English is 86\%:14\%. As shown in Figure \ref{fig:sft_data_distribution}, it comes from single-turn Chinese medical dialogues (single\_turn\_QA\_ch), single-turn English medical dialogues (single\_turn\_QA\_en), multi-turn Chinese medical dialogues (multi\_turn\_QA\_ch), and medical subject knowledge multiple-choice questions (single\_turn\_option\_ch).

\subsection{RLHF}
We enhance the model's capabilities using Direct Preference Optimization (DPO) \cite{rafailov2023direct} after the SFT stage. To align the model's output with human preferences while preserving the foundational abilities gained during the Continuous Pre-training and SFT stages \cite{lu2024online}, we construct subjective preference data and objective preference data. We also provide the training details of the DPO stage.

\subsubsection{Data Construction}
We construct the preference pair for the DPO stage using samples that have the same distribution as the SFT dataset. This mainly includes the following two preferences.

\noindent\textbf{Subjective Preference Data}
We aim to construct dpo pairs where the chosen response aligns closely with human preferences. For each prompt, we first ask GPT-4 to respond as a professional and helpful doctor. Then, using GPT-4, we evaluate the superiority or inferiority of the original response and this newly generated response from the prompt. The evaluation considers four aspects: Fluency, Relevance, Completeness, and Proficiency in Medicine \cite{zhang2023huatuogpt}. We select the superior response as the chosen response for the dpo pair and the inferior response as the rejection response.

\noindent\textbf{Objective Preference Data} While RLHF can guide LLMs to align with human expectations, numerous studies show that this method can cause LLMs to forget abilities acquired during pre-training and SFT stages \cite{bai2022training, dong2023abilities}, leading to an "alignment tax" \cite{dong2023raft,sun2024principle}. To mitigate this issue, we construct objective preference data. Specifically, for objective prompts with known ground truth answers, we consider the ground truth as the chosen response and randomly select incorrect answers from the remaining options as rejection responses. For instance, in multiple-choice questions, if the ground truth is option A, we randomly select from options B, C, and D to construct the rejection response.

\subsubsection{Training Details}

We constructed a dataset of 12,727 DPO preference pairs, consisting of 9,019 subjective and 3,708 objective data samples. We trained the model over two epochs using 8 NVIDIA Tesla A100 GPUs. The settings included a learning rate of 2e-7, a batch size of 64, and a beta of 0.03. Additionally, we employed a learning rate warmup and a cosine learning rate scheduler for optimization.





\section{Evaluation}

We evaluate our model's performance on several open-source Chinese and English benchmarks related to the medical domain. These benchmarks assess the model's ability to comprehend medical knowledge and engage in both single-turn and multi-turn conversations on medical topics.

\subsection{Medical Knowledge Benchmark}
We extract medical-related questions from the MMLU \cite{hendrycks2020measuring} and C-Eval \cite{huang2024c} benchmarks, and we utilize questions from the CMB-Exam \cite{wang2023cmb}, MedQA \cite{jin2021disease}, MedMCQA \cite{pal2022medmcqa} and PubMedQA \cite{jin2019pubmedqa} test set to evaluate the model's proficiency in medical knowledge.

\noindent\textbf{MMLU} is the english multi-subject multiple-choice dataset, from which we extract medical-related tasks to evaluate the model's performance. These tasks encompass various medical domains, including anatomy, clinical knowledge, college biology, college medicine, medical genetics, and professional medicine.

\noindent\textbf{C-Eval} is a chinese multiple-choice dataset. We extracted tasks related to medicine from the validation set, such as basic medicine, clinical medicine, medical practice, and veterinary medicine to test the model's performance.

\noindent\textbf{CMB-Exam} is a collection of multiple-choice questions in Chinese, sourced from various professional mdedical qualification examinations. It encompasses questions from exams for physicians, nurses, technicians, pharmacists, undergraduate medical programs, and graduate entrance examinations. We utilize 11,200 questions from the test set to conduct a comprehensive, multi-level assessment of the model's medical knowledge.

\noindent\textbf{MedQA} is a multiple-choice question dataset from the United States Medical Licensing Examination (USMLE). Its test set consists of 1,273 questions, which are used to assess a model's medical knowledge and reasoning skills required to obtain a medical license in the United States.

\noindent\textbf{MedMCQA} is a large-scale multiple-choice question and answer dataset, sourced from India's medical entrance exams (AIIMS/NEET). Its test set comprises 6,100 questions, enabling the evaluation of a model's general medical knowledge and reasoning abilities.

\noindent\textbf{PubMedQA} is a closed-domain question and answer dataset, where each question can be answered by referring to the relevant context from PubMed abstracts. We use 500 test questions from this dataset to evaluate a model's ability to understand and reason about biomedical literature.

\subsection{Medical Dialogue Benchmark}

\begin{table}[t]
\centering
\resizebox{0.48\textwidth}{!}{
\begin{tabular}{l|ccccc}
\hline 
Model & \textbf{MMLU} & \textbf{C-Eval} & \textbf{MedQA} & \textbf{MedMCQA} & \textbf{PubMedQA}\\ 

\hline
Aquila     & 42.91          & 48.77           & 38.65 & 38.58  & 71.60         \\
Aquila-Med & 49.32          & 48.40           & 41.56  & 38.23 & 72.40          \\ \hline
\end{tabular}
}
\caption{Performance on various medical knowledge benchmarks for continue pre-training. Specifically, MMLU and C-Eval represent the average scores obtained by the model on the medical-related sub-tasks within these benchmarks. Here, our setting is 3-shot.}
\label{tab:cpt_know}
\vspace{-0.5cm}
\end{table}

\begin{table}[t]
\centering
\resizebox{0.48\textwidth}{!}{
\begin{tabular}{l|ccc}
\hline
                    & \textbf{MMLU} & \textbf{C-Eval} & \textbf{CMB-Exam} \\ \hline
Aquila-Med-Chat     & 56.2          & 50.44           & 47.63             \\
Aquila-Med-chat (RL) & 56.4          & 53.10           & 47.12             \\ \hline
\end{tabular}
}

\caption{Performance on various medical knowledge benchmarks for supervised fine-tuning. Specifically, MMLU and C-Eval represent the average scores obtained by the model on the medical-related sub-tasks within these benchmarks.}
\label{tab:sft_know}
\end{table}

 \begin{figure}[t]
 \centering
\resizebox{0.48\textwidth}{!}{
 \includegraphics{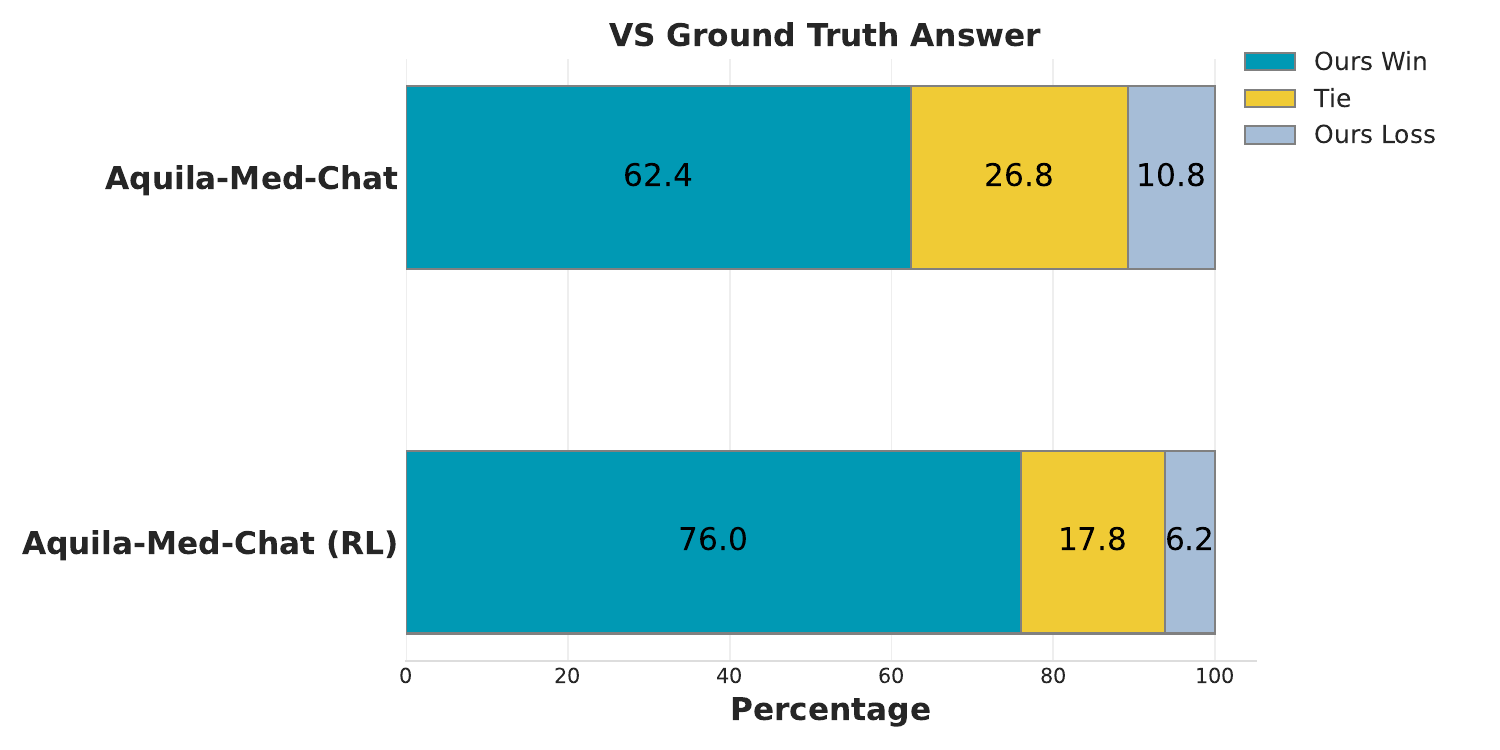}
 }
 \caption{The comparison of our model's predicted answers and the ground truth answers from the dataset on single-round dialogues from the Huatuo MedicalQA.}
 \vspace{-0.5cm}
 \label{fig:vs_ground}

\end{figure}

\begin{figure*}[t]
    \centering
    \begin{subfigure}[b]{0.24\textwidth}
        \centering
        \includegraphics[width=\textwidth]{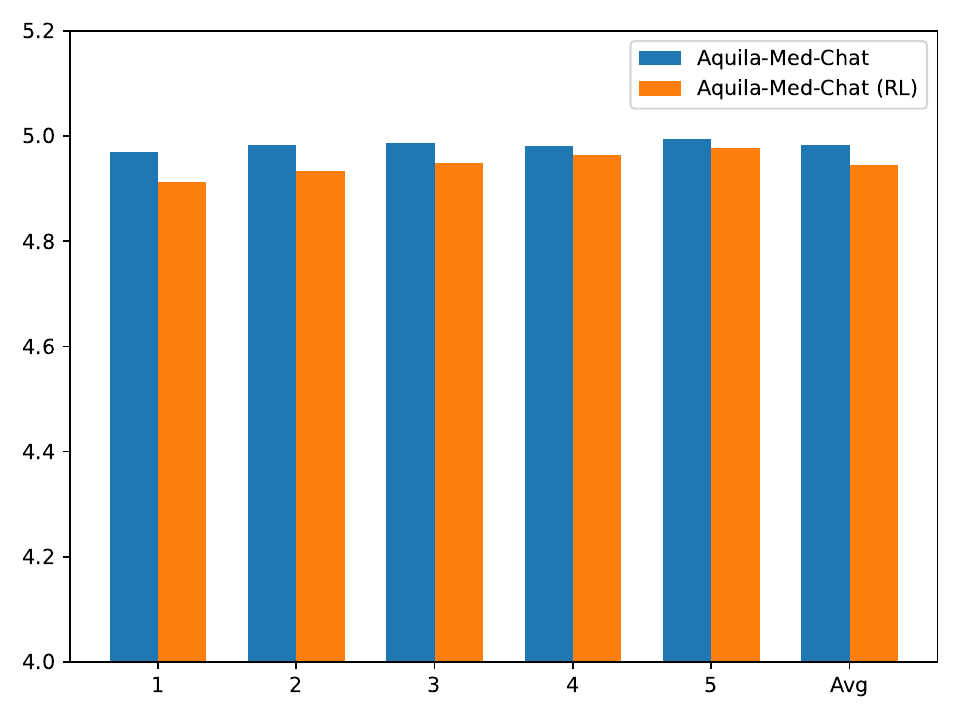}
        \caption{Fluency Score}
    \end{subfigure}
    \hfill
    \begin{subfigure}[b]{0.24\textwidth}
        \centering
        \includegraphics[width=\textwidth]{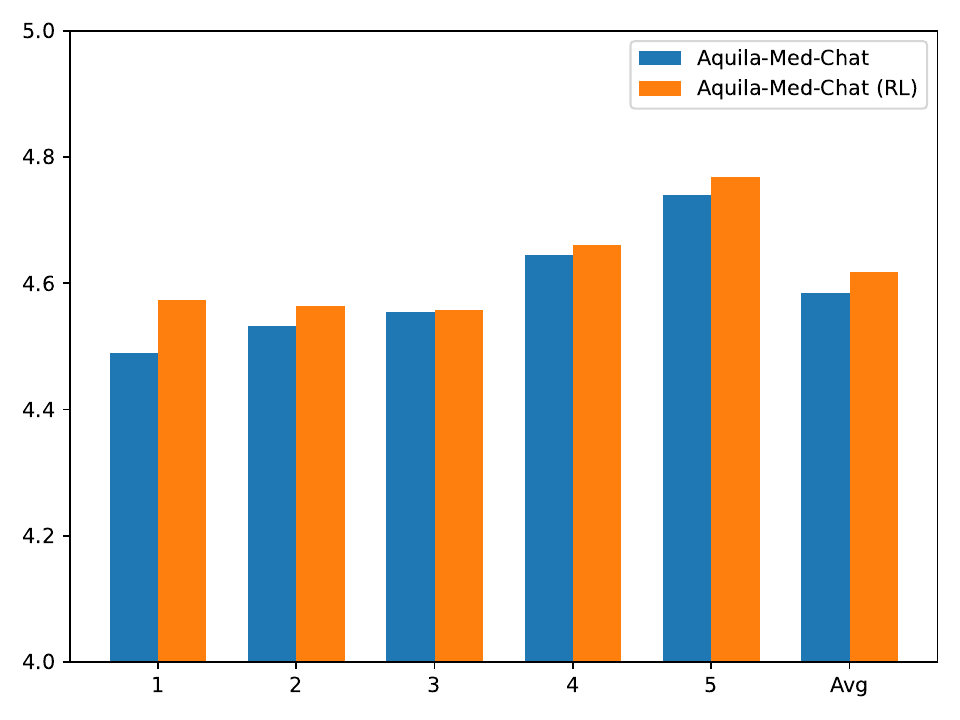}
        \caption{Relevance Score}
    \end{subfigure}
    \hfill
    \begin{subfigure}[b]{0.24\textwidth}
        \centering
        \includegraphics[width=\textwidth]{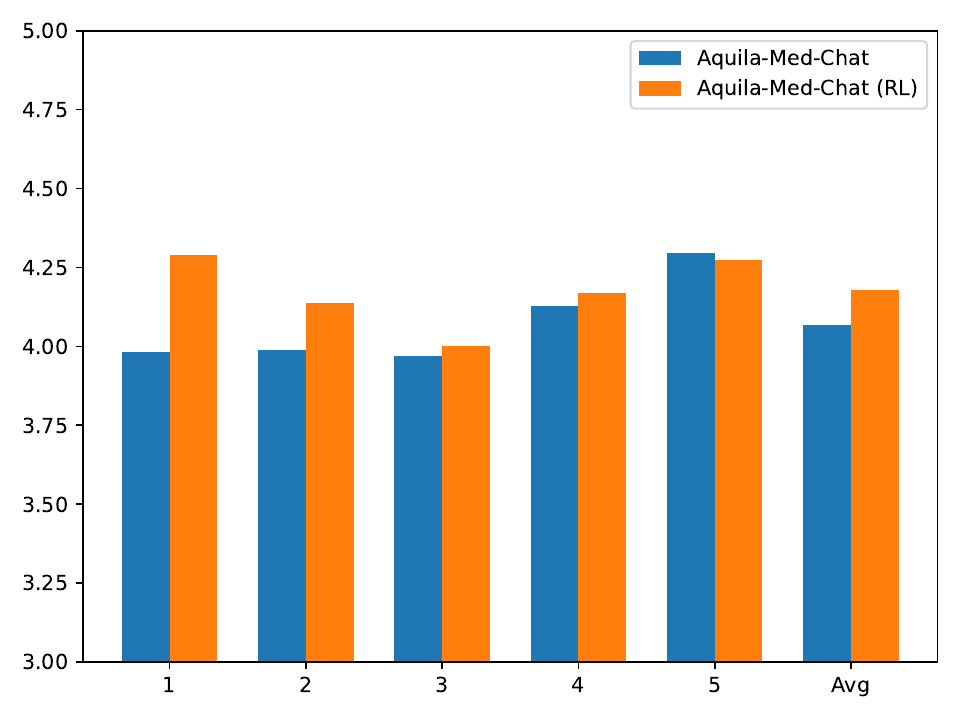}
        \caption{Completeness Score}
    \end{subfigure}
    \hfill
    \begin{subfigure}[b]{0.24\textwidth}
        \centering
        \includegraphics[width=\textwidth]{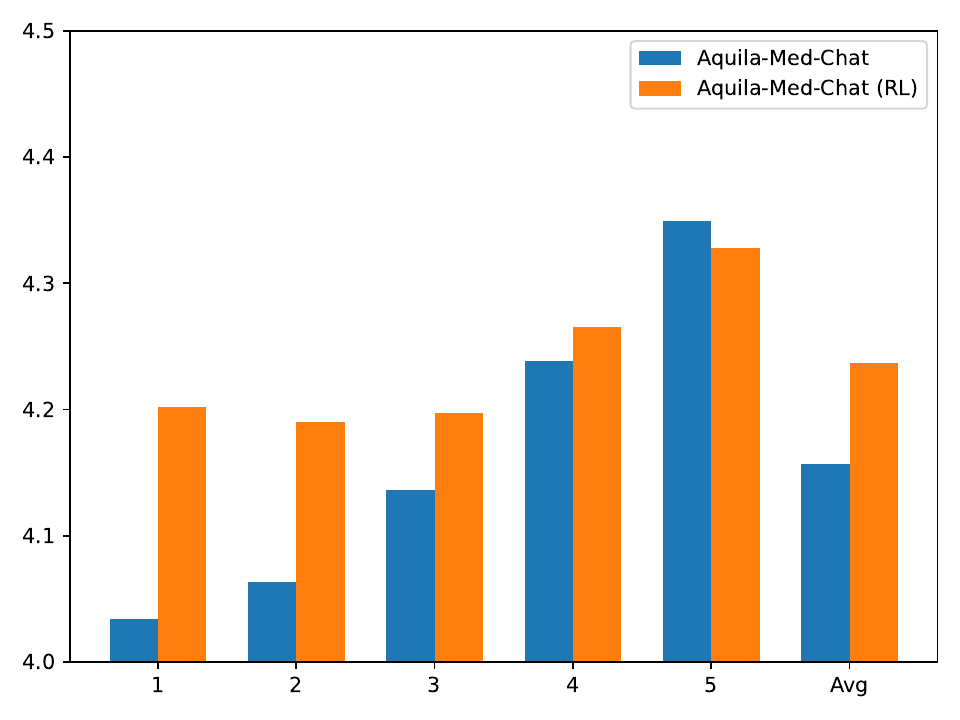}
        \caption{Proficiency Score}
    \end{subfigure}
    \caption{Performance on the CMTMedQA dataset in multi-round dialogues. The x-axis represents different rounds of the dialogue, while the "Avg" data point displays the average score across all rounds.}
    \label{fig:multi_turn_CMTMedQA}

\end{figure*}

\begin{figure*}[t]
    \centering
    \begin{subfigure}[b]{0.24\textwidth}
        \centering
        \includegraphics[width=\textwidth]{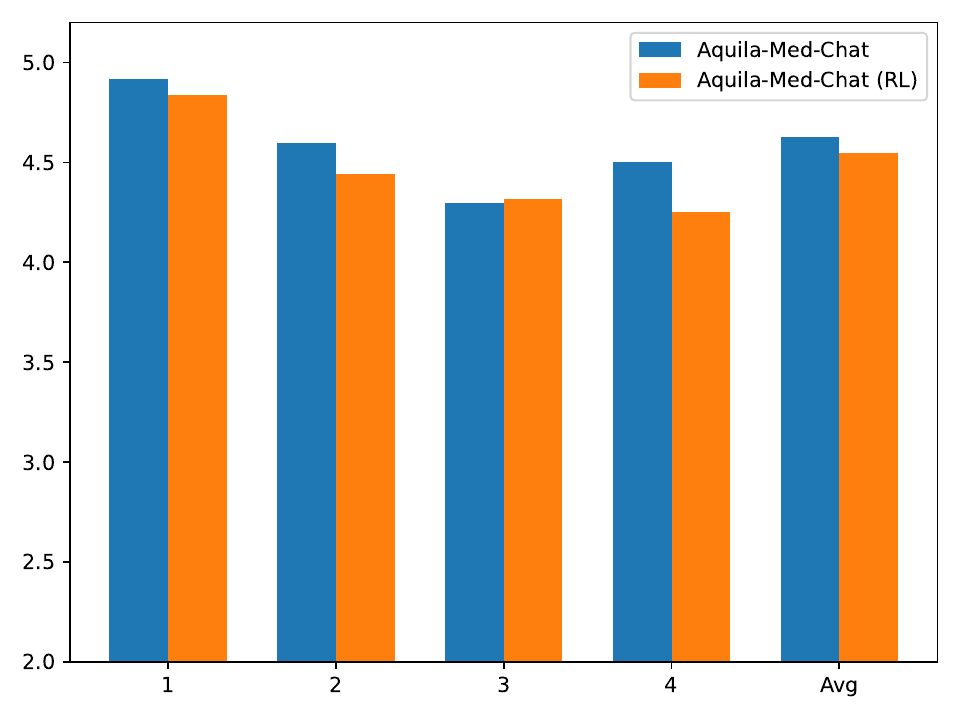}
        \caption{Fluency Score}
    \end{subfigure}
    \hfill
    \begin{subfigure}[b]{0.24\textwidth}
        \centering
        \includegraphics[width=\textwidth]{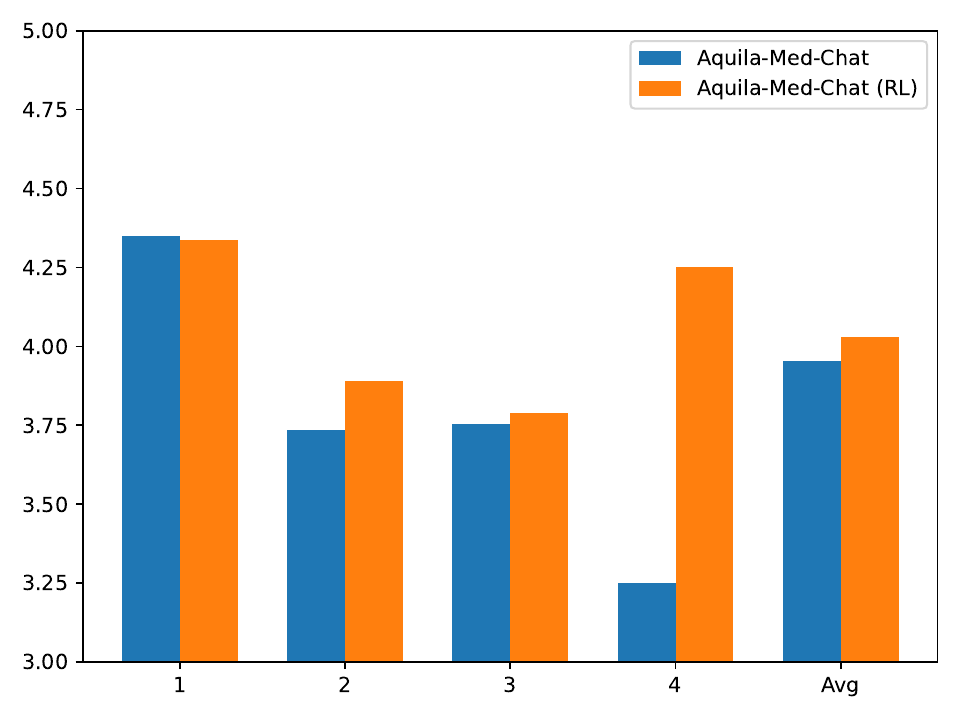}
        \caption{Relevance Score}
    \end{subfigure}
    \hfill
    \begin{subfigure}[b]{0.24\textwidth}
        \centering
        \includegraphics[width=\textwidth]{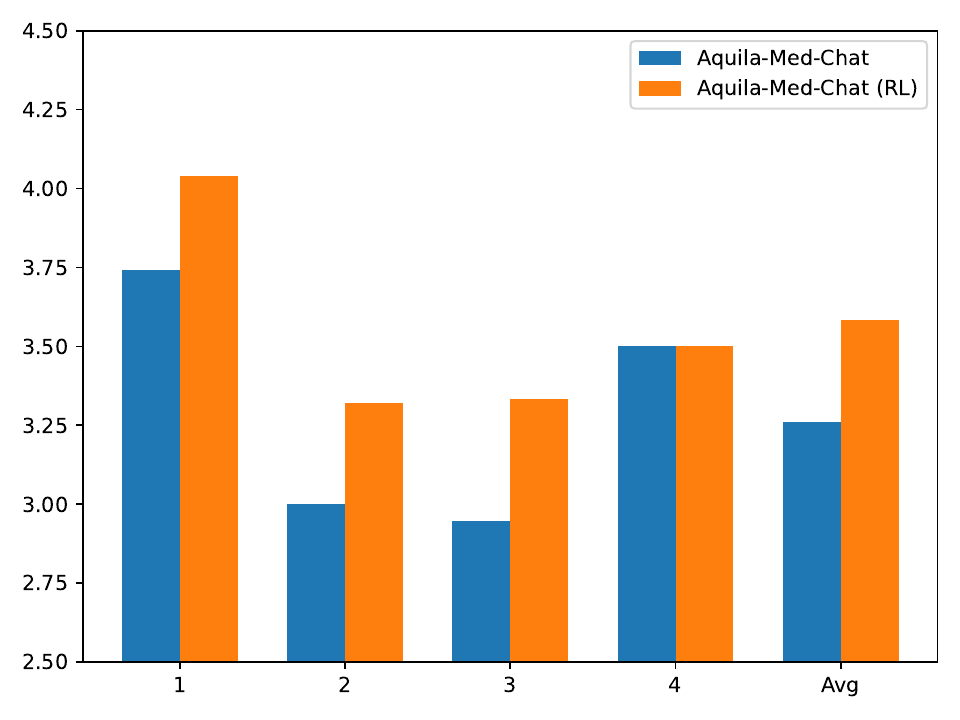}
        \caption{Completeness Score}
    \end{subfigure}
    \hfill
    \begin{subfigure}[b]{0.24\textwidth}
        \centering
        \includegraphics[width=\textwidth]{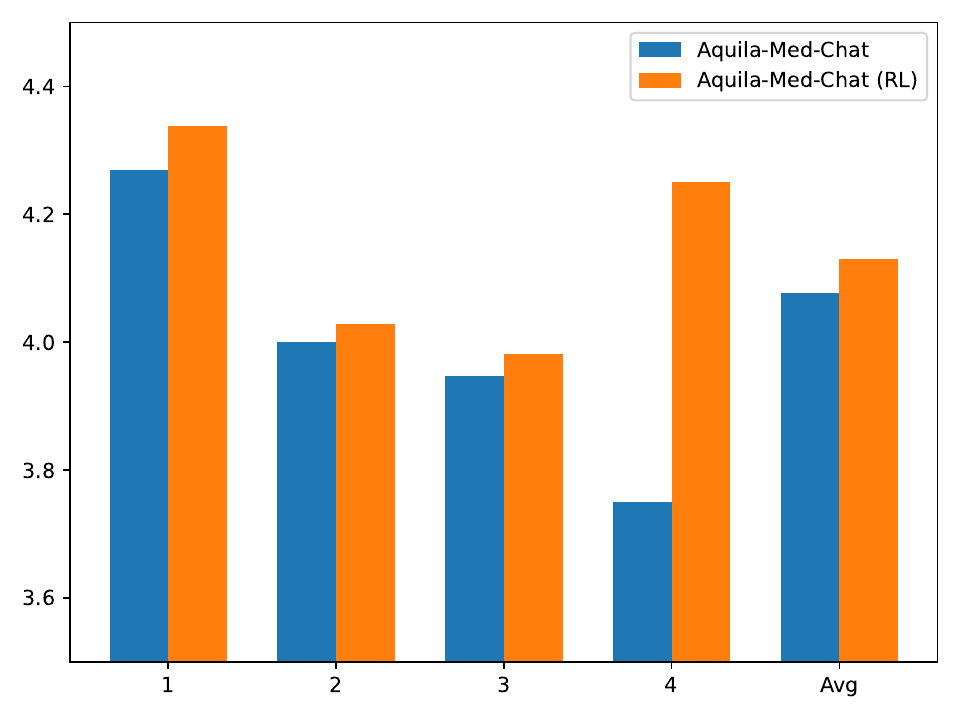}
        \caption{Proficiency Score}
    \end{subfigure}
    \caption{
    Performance on the CMT-Clin dataset in multi-round dialogues. The x-axis represents different rounds of the dialogue, while the "Avg" data point displays the average score across all rounds.}
    \label{fig:multi_turn_Clin}

\end{figure*}

 \begin{figure}[t]
 \centering
\resizebox{0.48\textwidth}{!}{
 \includegraphics{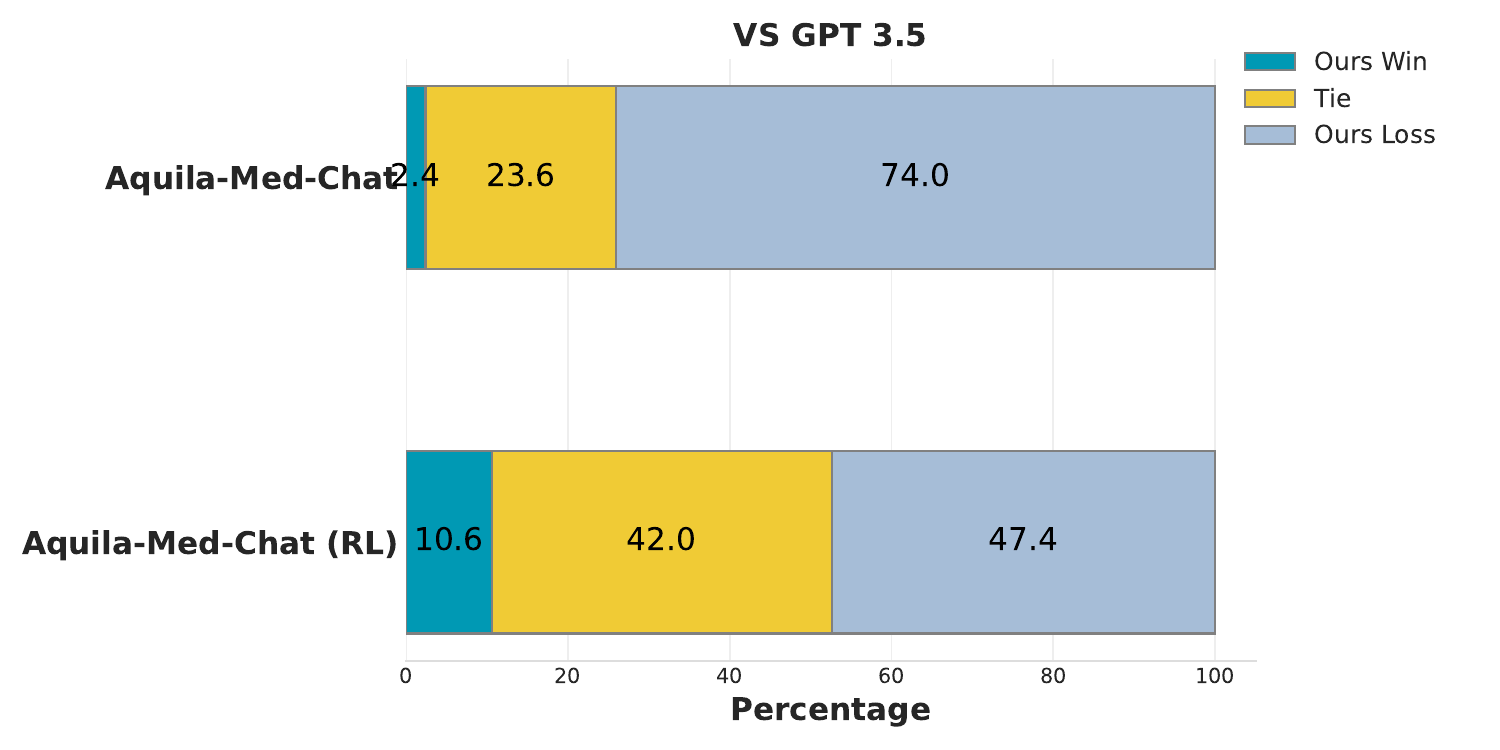}
 }
 \caption{The comparison of our model's predicted answers and the gpt-3.5 predicted answers on single-round dialogues from the Huatuo MedicalQA.}
 \vspace{-0.5cm}
 \label{fig:vs_gpt}

\end{figure}

We evaluate the model's capability to solve realistic patient problems by assessing its medical knowledge and complex reasoning abilities. This evaluation covers single-round dialogue scenarios, such as the Huatuo MedicalQA \cite{li2023huatuo26m}, as well as multi-round dialogue scenarios like CMtMedQA \cite{yang2023zhongjing} and CMB-Clin \cite{wang2023cmb}.

\noindent\textbf{Huatuo MedicalQA} is a large-scale Chinese Medical Question Answering (QA) dataset, and we use its test set to evaluate the model's capability in single-round dialogues. Specifically, we sample 500 question-answer pairs from the test set and employ GPT-4 to compare the model's predicted answers with other reference answers (mainly including the ground truth answer from the dataset and the answer generated by GPT-3.5). Inspired by \citet{zhang2023huatuogpt}, we use the prompt in Table \ref{tab:single_judge} to judge the quality of the answers. Considering that GPT-4 may exhibit a "position bias" when judging \cite{zheng2024judging}, we swap the order of the predicted answer and the reference answer. We determine a answer as winning or losing only when the judgment results are completely consistent before and after the swap.

\noindent\textbf{CMtMedQA} is a large-scale dataset consisting of multi-turn medical dialogues in Chinese. To evaluate the model's ability to engage in complex dialogues and initiate proactive inquiries, we utilized approximately 1,000 samples from the dataset's test set.

\noindent\textbf{CMB-Clin} consists of 74 expertly curated medical case consultations derived from clinical diagnostic teaching materials. It evaluates the model's mastery and reasoning abilities in applying medical knowledge through multi-round diagnostic dialogues.

For multi-round dialogue datasets such as CMtMedQA and CMB-Clin, inspired by \citet{wang2023cmb}, we employed GPT-4 to evaluate the model's responses in each round of the dialogue. The evaluation focused on four key aspects: fluency, relevance, completeness, and proficiency in medical knowledge. The specific evaluation prompt used is displayed in Table \ref{tab:multi_judge}.


\section{Experimental Results}

\subsection{Results for Continue Pre-training}
Table \ref{tab:cpt_know} shows the results of our continue pre-training on five benchmarks. It can be observed that Aquila-Med has improved to a certain extent compared with Aquila, especially on MMLU. This shows that even if the model uses the data which has been already learned in the pre-training stage, the professional ability of the model can be further improved by improving the quality and professional density. In general, we obtain a basic model with medical domain knowledge.

\subsection{Results for Alignment}
For instruct-tuning, we evaluate it from two aspects: medical subject questions and doctor-patient consultation. Table \ref{tab:sft_know} shows the results on three medical knowledge benchmarks. We found that Aquila-Med-Chat has good command following ability, and Aquila-Med-Chat (RL) has made further progress, especially C-Eval. Figures \ref{fig:vs_ground} and Figure \ref{fig:vs_gpt} show the comparison of the outputs of our models with the reference and GPT-3.5 outputs in single-turn dialogues. It is observed that both Aquila-Med-Chat and Aquila-Med-Chat (RL) have achieved good results, especially Aquila-Med-Chat (RL) has achieved human-style alignment. For multi-turn dialogues, we use GPT-4 to score each turn in four dimensions, and the results are shown in Figure \ref{fig:multi_turn_CMTMedQA} and Figure \ref{fig:multi_turn_Clin}. The evaluation results indicate that Aquila-Med-Chat (RL) performed well in terms of generating fluent responses. Additionally, it was observed that Aquila-Med-Chat (RL) significantly enhanced the model's performance in terms of relevance, completeness, and proficiency, while still maintaining a high level of fluency in the generated responses.

\section{Conclusion}
In this paper, we present Aquila-Med, a bilingual medical LLM designed to address the challenges of specialized medical knowledge through continued pre-training, SFT, and RLHF. Our extensive dataset construction and training process have led to significant improvements in the model's ability to handle single-turn and multi-turn medical consultations, as well as medical multiple-choice questions. Aquila-Med's strong performance on various benchmarks validates the effectiveness of our approach. By open-sourcing our datasets and training processes, we aim to facilitate further advancements in the development of medical LLMs within the research community.

\bibliography{anthology,custom}
\bibliographystyle{acl_natbib}

\clearpage

\appendix

\section{Prompt For Judging}

\begin{table}[t]
\centering
\resizebox{0.48\textwidth}{!}{
\begin{tabular}{l}
\hline
\begin{tabular}[c]{@{}l@{}}{[}User{]}\\ \{user\_query\}\\ {[}End of User{]}\\ {[}Assistant 1{]}\\ \{assistant1\}\\ {[}End of Assistant 1{]}\\ {[}Assistant 2{]}\\ \{assistant2\}\\ {[}End of Assistant 2{]}\\ {[}System{]}\\ We would like to request your feedback on two multi-turn conversations \\ between the AI assistant and the user displayed above. Requirements: \\ Focus on the AI’s response in the conversation. The AI assistant should \\ act like the doctor using the tone, manner, and vocabulary the human \\ doctor would use. It should be to the point, without unnecessary \\ elaboration or extraneous information. The AI assistant should respond \\ appropriately to the user in a manner that helps to progress the \\ conversation. The description of symptoms should be comprehensive \\ and accurate, and the provided diagnosis should be the most reasonable \\ inference based on all relevant factors and possibilities. The treatment \\ recommendations should be effective and reliable, taking into account \\ the severity or stages of the illness. The prescriptions should be effective \\ and reliable, considering indications, contraindications, and dosages. \\ Please compare the performance of the AI assistant in each conversation. \\ You should tell me whether Assistant 1 is ‘better than‘, ‘worse than‘, or \\ ‘equal to‘ Assistant 2. Please first compare their responses and analyze \\ which one is more in line with the given requirements.\\ \\ In the last line, please output a single line containing only a single label \\ selecting from 'Assistant 1 is better than Assistant 2', 'Assistant 1 is worse \\ than Assistant 2', and 'Assistant 1 is equal to Assistant 2'.\end{tabular} \\ \hline
\end{tabular}
}
\vspace{-10pt}
\caption{Prompt for judging the quality of a single-round dialogue}
\vspace{-15pt}
\label{tab:single_judge}
\end{table}

\begin{table}[t]
\centering
\resizebox{0.48\textwidth}{!}{
\begin{tabular}{l}
\hline
\begin{tabular}[c]{@{}l@{}}You are an AI evaluator specializing in assessing the quality of answers\\ provided by other language models . Your primary goal is to rate the\\ answers based on their fluency , relevance , completeness , proficiency\\ in medicine . Use the following scales to evaluate each criterion :\\ Fluency :\\ 1: Completely broken and unreadable sentence pieces\\ 2: Mostly broken with few readable tokens\\ 3: Moderately fluent but with limited vocabulary\\ 4: Mostly coherent in expressing complex subjects\\ 5: Human - level fluency\\ Relevance :\\ 1: Completely unrelated to the question\\ 2: Some relation to the question , but mostly off - topic\\ 3: Relevant , but lacking focus or key details\\ 4: Highly relevant , addressing the main aspects of the question\\ 5: Directly relevant and precisely targeted to the question\\ Completeness :\\ 1: Extremely incomplete\\ 2: Almost incomplete with limited information\\ 3: Moderate completeness with some information\\ 4: Mostly complete with most of the information displayed\\ 5: Fully complete with all information presented\\ Proficiency in medicine :\\ 1: Using plain languages with no medical terminology .\\ 2: Equipped with some medical knowledge but lacking in - depth details\\ 3: Conveying moderately complex medical information with clarity\\ 4: Showing solid grasp of medical terminology but having some minor\\ mistakes in detail\\ 5: Fully correct in all presented medical knowledge\\ You will be provided with the following information :\\ - a conversation\\ - a question based on the conversation\\ - the solution to the question\\ - a model ’ s answer to the question\\ {[} conversation {]}\\ \{history\}\\ {[} end of conversation {]}\\ {[} question {]}\\ \{question\}\\ {[} end of question {]}\\ {[} solution {]}\\ \{solution\}\\ {[} end of solution {]}\\ {[} answer {]}\\ \{answer\}\\ {[} end of answer {]}\\ Make sure to provide your evaluation results in JSON format and ONLY the\\ JSON , with separate ratings for each of the mentioned criteria as in\\ the following example :\\ \{"fluency": 3, "relevance": 3, "completeness": 3, "proficiency": 3\}\end{tabular} \\ \hline
\end{tabular}
}
\vspace{-10pt}
\caption{Prompt for judging the quality of a multi-round dialogue}
\vspace{-15pt}
\label{tab:multi_judge}
\end{table}


\end{document}